\documentclass[10pt, a4paper]{article}
\usepackage{lrec2000}
\usepackage{alltt}
\usepackage{epsfig, url}

\def\myurl#1{{\small [\url{#1}]}}
\def\myurlf#1{{\scriptsize [\url{#1}]}}
\def\smtt#1{{\footnotesize\texttt{#1}}}
\newenvironment{sv}{\scriptsize\begin{alltt}}{\end{alltt}\normalsize}

\title{Creating Annotation Tools with the Annotation Graph Toolkit}

\name{
Kazuaki Maeda,
Steven Bird,
Xiaoyi Ma, and
Haejoong Lee
}

\address{
  Linguistic Data Consortium, University of Pennsylvania\\
  3615 Market Street, Philadelphia, PA 19104-2608, USA\\
\{maeda, sb, xma, haejoong\}@ldc.upenn.edu
}

\abstract{
  The Annotation Graph Toolkit is a collection of software supporting
  the development of annotation tools based on the annotation graph
  model.  The toolkit includes application programming interfaces for
  manipulating annotation graph data and for importing data from other
  formats.  There are interfaces for the scripting languages Tcl and
  Python, a database interface, specialized graphical user interfaces
  for a variety of annotation tasks, and several sample applications.
  This paper describes all the toolkit components for the benefit of
  would-be application developers.  }

\begin{document}

\maketitleabstract

\section{Introduction}

\begin{figure}[htbp]
  \begin{center}
    \leavevmode
  \epsfig{figure=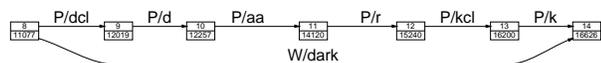,width=0.95\linewidth}
  \end{center}
\caption{Annotation Graph for a fragment of TIMIT data}
\label{fig:ag-timit2}
\end{figure}

Linguistic databases are widely used in the scientific study of language
and in language-technology research and development.  Many software tools
have been developed to support the creation of annotated linguistic
databases, and some of them are documented in \cite{BirdHarrington01}.
Bird and Liberman have developed a model for expressing the logical
structure of linguistic annotations, and have demonstrated that it can
encode a great variety of existing annotation types \cite{BirdLiberman01}.
An annotation graph is a directed acyclic graph where edges are labeled
with fielded records, and nodes are (optionally) labeled with time offsets.
Figure~\ref{fig:ag-timit2} shows an annotation graph for a fragment of the
TIMIT corpus \cite{TIMIT86}.

Annotation graphs have opened up new possibilities for creation,
maintenance and search, and lead to new annotation tools with applicability
across the text, audio and video modalities.  Annotation graphs are also
permitting existing annotation tools -- each with large user-bases -- to be
made fully interoperable.

The Annotation Graph Toolkit (AGTK) is a collection of software
supporting the development of annotation tools based on the annotation graph
model.  AGTK includes application programming interfaces
for manipulating annotation graph data and for importing data from other
formats, a database interface, wrappers for scripting languages,
specialized graphical user interfaces for annotation tasks, and sample
applications.

This paper is addressed to developers of linguistic annotation tools.
We begin with a high-level overview of the tool architecture
(\S 2), before presenting the most important features
of the annotation graph library (\S 3), the file I/O library
(\S 4), and their scripting language interfaces
(\S 5).  In \S 6 we discuss our approach to
tool creation, involving rapid high-level programming in scripting
languages, interfaced to stable and optimized C++ libraries.  We describe
our model of inter-component communication, and
our approach to GUI design, in which we create many special-purpose tools
that are maximally ergonomic for the task at hand.  All
software, interface definitions, configuration files and data samples are
available under an open source license.

\section{Architecture}
\label{sec:architecture}

Existing annotation tools are based on a two level model.  The system we
discuss in this paper is based on a three level model, in which annotation
graphs provide a logical level independent of application and physical
levels.  This is the three-level model of modern database systems
\cite{Abiteboul95} applied to linguistic databases in support of data
independence, data reuse, and software integration.  The application level
represents special-purpose tools built on top of the general-purpose
infrastructure at the logical level.

\begin{figure}[htbp]
\centerline{\epsfig{file=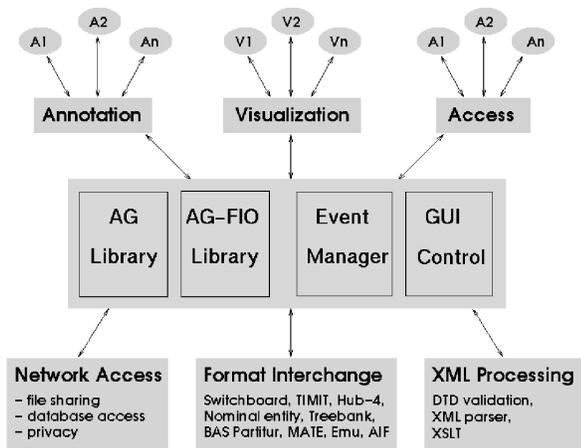,width=0.95\linewidth}}
\caption{Architecture for Annotation Systems}
\label{fig:arch}
\end{figure}

The toolkit is comprised of several components, structured according
to the architecture shown in Figure~\ref{fig:arch}.  This model
permits applications to abstract away from file format issues, and
deal with annotations purely at the logical level, through the
annotation graph API.  Annotation tools provide graphical user
interface components both for signal visualization and for annotation,
and the communication between these components is handled by an
extensible event language.

As with other recent architectures for language technologies, e.g. \cite{Allen00},
the architecture consists of a set of loosely-coupled, heterogenous
components that communicate with each other by exchanging messages.
This design has three benefits.
First, components can be implemented in the most opportune language,
and wrappers can easily be added to legacy and third-party components.
Second, message traffic can be logged to facilitate error diagnosis
and to permit inter-component and human-computer interactions to be replayed
and analyzed.
Third, message passing permits the transport
protocol to be separated from the communication content.  The former is
enforced by the infrastructure, while the latter is extremely flexible.

\section{The Annotation Graph Library}
\label{sec:aglib}

\begin{figure}[htbp]
  \begin{center}
    \leavevmode
    \epsfig{figure=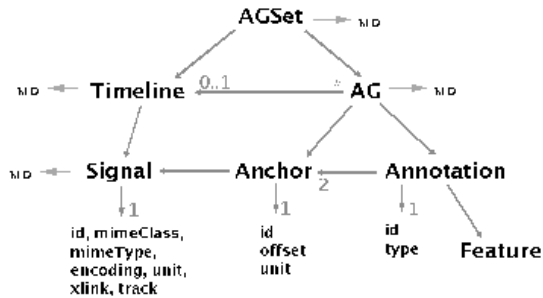, width=0.95\linewidth}
  \end{center}

  \caption{The AG Object Model}
  \label{fig:object-model}
\end{figure}

The annotation graph library (libag) is implemented in C++,\footnote{The library has
  recently been ported to Java.} and provides functions for creating,
deleting, modifying and searching the following annotation graph objects: \emph{AGSet},
\emph{AG}, \emph{Annotation}, \emph{Anchor}, \emph{Timeline},
\emph{Signal}, \emph{Feature} and \emph{Metadata}.  These objects are
related to each other according to the object model shown in
Figure~\ref{fig:object-model}.  The various objects will be explained in
more detail as we describe the API.

The annotation graph library also keeps indexes for the \emph{Annotation}, \emph{Anchor},
\emph{Feature} and \emph{Metadata} types so that searches can be done
efficiently.

The API provides access to internal objects -- signals, anchors, annotations,
etc -- through the use of identifier strings.  The types of annotation graph
identifiers include \smtt{AGSetId},
\smtt{AGId}, and \smtt{AnnotationId}.


\subsection{The structure of annotation graph identifiers}

All identifiers are represented as strings.
The internal structure of an identifier can best be understood
in terms of the object hierarchy in Figure~\ref{fig:object-model}.
Annotation graph identifiers are \emph{fully-qualified}: given an object identifier,
all of the ancestor objects can be discovered simply by inspecting
the identifier.  For instance, an anchor \smtt{"Timit:AG1:Anchor2"}
belongs to the annotation graph \smtt{"Timit:AG1"}, which in turn belongs to the
AGSet \smtt{"Timit"}.

Internally, the annotation graph library maps these string identifiers
to object references. A consequence of this design is that annotation
graph objects can be referenced from scripting languages using
human-readable names.  The use of fully-qualified identifiers also
reduces the risk of collision -- the accidental re-use of the same
identifier in different places -- which can have unpredictable
consequences.

\subsection{Annotation graph API functions}

This section explains some typical API functions.  The complete IDL
definition of the AG-API is available online
\myurl{http://www.ldc.upenn.edu/AG/}.

\subsubsection{AGSet and AG functions}

An AGSet is an object which contains a set of annotation graphs.
Typically, an AGSet corresponds to a corpus, but it might also
correspond to a user-specified selection from a corpus, or to a
selection spanning several different corpora.  The first thing to do
in working with annotation graphs is to create an AGSet object to hold
them.

\noindent
\textbf{CreateAGSet.} This creates an empty AGSet with a specified
 AGSetId, and returns the AGSetId:

\begin{sv}
AGSetId CreateAGSet(AGSetId agSetId);
\end{sv}

Once an AGSet is created, Timelines, Signals, AGs, and then Annotations
and Anchors can also be created.  Certain functions can then be called
on these data types, for example, to test for their existence or to
delete them.
An AGSet can be deleted by using \smtt{DeleteAGSet}. Its existence can
be tested by using \smtt{ExistsAGSet}.

\noindent
\textbf{CreateAG.} This creates an AG and returns the AGId; it throws an
AGException if the \texttt{id} does not contain a valid AGSetId, or if
the timeline does not exist:

\begin{sv}
AGSet CreateAG(Id id);
AGSet CreateAG(Id id, TimelineId timelineId);
\end{sv}

The parameter \texttt{id} may be either an AGSetId or an AGId. If it is an
AGSetId, an AGId will be assigned to the new AG.  However, if it is an
AGId, the library will try to use the supplied id.  If this id is
unavailable, it will assign a new AGId.

The timelineId is the id of the timeline with which the new AG will be
associated. An AG can be created without being associated to any timeline.

\subsubsection{Timeline and Signal functions}

A ``timeline'' is a collection of synchronized signals, such as
separate audio and video recordings of the same event, or a
multichannel recording of a conference call.  The key defining
property of a timeline is that the offsets into its signals are
intertranslatable; any offsets into any one of the associated signals
can be mapped to an offset into any of the others.  Whenever multiple
signals or channels are used in an annotation task, they are assigned
to a distinct \smtt{Signal} object.\footnote{
  An exception to this is the situation of a multichannel recording in
  which the different tracks are not discriminated in the annotation task;
  here they may be treated as a single \smtt{Signal} object.}
Multiple synchronized signals are
grouped into a single \smtt{Timeline} object.

The functions \smtt{CreateTimeline}, \smtt{ExistsTimeline} and 
\smtt{DeleteTimeline} will now be explained.
\vspace{1ex}

\noindent
\textbf{CreateTimeline.} This creates a new \emph{Timeline} and returns
the \smtt{TimelineId}:

\begin{sv}
TimelineId   CreateTimeline(Id id);
\end{sv}

\smtt{Id} is \smtt{AGSetId} or \smtt{TimelineId}. In either case, the
\emph{AGSet} to which the \emph{Timeline} belongs must already exist;
otherwise an exception will be thrown. For example:

\begin{sv}
/* Create an AGSet with id "Timit" */
AGSetId agSetId = CreateAGSet("Timit");

/* Create a new timeline */
TimelineId timeline1 = CreateTimeline(agSetId); 

/* Create another timeline */ 
TimelineId timeline2 =
             CreateTimeline("Timit:Timeline2");

/* The following causes an exception
since AGSet "CallHome" does not exist */
TimelineId timeline3 =
             CreateTimeline("CallHome");

/* The following also causes an exception */
TimelineId timeline4 =
             CreateTimeline("CallHome:Timeline2");
\end{sv}

\noindent
\textbf{ExistsTimeline.} This tests for the existence of the specified
\emph{Timeline}, and returns true if it exists and false otherwise:

\begin{sv}
boolean ExistsTimeline(TimelineId timelineId);
\end{sv}

\noindent
\textbf{DeleteTimeline.} This deletes the specified \emph{Timeline} if it
exists:

\begin{sv}
void DeleteTimeline(TimelineId timelineId);
\end{sv}

\noindent
\textbf{CreateSignal.} This creates a new signal and adds it to the timeline.  

\begin{sv}
SignalId CreateSignal(Id id, URI uri, 
  MimeClass mimeClass, MimeType mimeType, 
  Encoding encoding, Unit unit, Track track);
\end{sv}

The \texttt{id} argument might be TimelineId or SignalId. If it is a
TimelineId, the library will generate a new SignalId. If it is a SignalId,
the library will try the given id first, and if it's taken, generate a new
SignalId. If the id given is invalid, it throws an AGException.

The \texttt{uri} argument specifies a location where the signal is to be
found.  Applications may use this information to display and replay a
signal.  The \texttt{mimeClass} and \texttt{mimeType} arguments tell an
application about the format of the signal, while the \texttt{encoding}
argument specifies how samples are coded (e.g. mu-law).  The \texttt{unit}
argument specifies the sample rate of the signal; annotation applications
may use this information to set the granularity of time coding and time
alignment in a user interface.  The \texttt{track} argument records which
track of the signal file contains the signal.  In this way, we can create
two or more distinct \smtt{Signal} objects which reference different tracks of the
same signal file.

\noindent
\textbf{GetSignals.} This returns \smtt{SignalIds} of the \emph{Signals}
contained in the specified \emph{Timeline}:

\begin{sv}
SignalIds GetSignals(TimelineId timelineId);
\end{sv}

The \smtt{SignalIds} are separated by spaces. 

\subsubsection{Annotation Functions}

\noindent
\textbf{CreateAnnotation.} This creates a new annotation:

\begin{sv}
AnnotationId CreateAnnotation(Id id,
                   AnchorId start, AnchorId end,
                   AnnotationType annotationType);
\end{sv}

The \texttt{id} argument can be an AGId or an AnnotationId.  If it is an
AGId, an AnnotationId will be assigned to the new annotation.  On the other
hand, if it is an AnnotationId, the library will try to use the supplied
id.  If this id is unavailable, it will assign a new AnnotationId.
The other arguments are as follows:
\emph{start} is the id of the start anchor; \emph{end} is the id of the
end anchor; \emph{annotationType} is the type of the annotation.

\smtt{CreateAnnotation} returns the AnnotationId of the new annotation.

\noindent
\textbf{ExistsAnnotation.} This returns true if the annotation exists:

\begin{sv}
bool ExistsAnnotation(AnnotationId annotationId);
\end{sv}

\noindent
\textbf{DeleteAnnotation.} This deletes an annotation:

\begin{sv}
void DeleteAnnotation(AnnotationId annotationId);
\end{sv}

\noindent
\textbf{CopyAnnotation.} This copies an existing annotation, with a new
identifier assigned to the new annotation:

\begin{sv}
AnnotationId CopyAnnotation(AnnotationId annotationId);
\end{sv}

\noindent
\textbf{SplitAnnotation.} This splits an annotation into two, creating a
new annotation with the same label data as the original one; returns
\smtt{Ids} of both annotations.

\begin{sv}
AnnotationIds SplitAnnotation(AnnotationId annotationId);
\end{sv}

\begin{figure}[htbp]
  \begin{center}
    \leavevmode
    \epsfig{figure=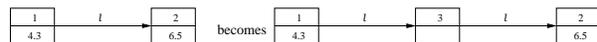,width=0.95\linewidth}  
  \end{center}
  \caption{Split an annotation}
\end{figure}

\noindent
\textbf{NSplitAnnotation.} This splits an annotation into N annotations,
creating N-1 new annotations having the same label data as the original
one; returns \texttt{id}s of all annotations, including the original one:

\begin{sv}
AnnotationIds NSplitAnnotation(
                AnnotationId annotationId, short N);
\end{sv}

\begin{figure}[htbp]
  \begin{center}
    \leavevmode
    \epsfig{figure=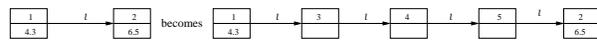,width=0.95\linewidth}  
  \end{center}
  \caption{Nsplit an annotation with N = 4}
\end{figure}

\subsubsection{Accessing Label Data}

\noindent
\textbf{SetFeature.} This sets the features of an annotation as well as the
features of the metadata associated with \emph{AGSets}, \emph{AGs},
\emph{Timelines} and \emph{Signals}.

\begin{sv}
void SetFeature(Id id, FeatureName featureName,
                      FeatureValue featureValue);
\end{sv}

The \smtt{Id} can be \smtt{AnnotationId}, \smtt{AGSetId}, \smtt{AGId},
\smtt{TimelineId} or \smtt{SignalId}. This is also true for other
\emph{Feature} functions, such as \smtt{ExistsFeature},
\smtt{DeleteFeature}, \smtt{GetFeature}, etc.
  
\noindent
\textbf{GetAnchorSet.} This returns all the \emph{Anchors} in a given \emph{AG}.

\begin{sv}
AnchorIds GetAnchorSet(AGId agId)
\end{sv}

\noindent
\textbf{GetAnchorSetByOffset.} This returns all anchors with its offset in
between offset-epsilon and offset+epsilon, inclusive. The default value
for epsilon is 0.

\begin{sv}
AnchorIds GetAnchorSetByOffset(AGId agId,
            Offset offset, float epsilon=0);
\end{sv}

\noindent
\textbf{GetAnchorSetNearestOffset.} This returns all anchors at the
nearest offset to the given offset:

\begin{sv}
AnchorIds GetAnchorSetNearestOffset(
            AGId agId, Offset offset);
\end{sv}

\subsubsection{Accessing Annotations}

\noindent
\textbf{GetIncomingAnnotationSet.} This returns the incoming annotations
of the specified anchor.  The incoming annotations of anchor
\smtt{a} are the annotations which end with anchor \smtt{a}:

\begin{sv}
AnnotationIds GetIncomingAnnotationSet(
                AnchorId anchorId);
\end{sv}

\noindent
\textbf{GetOutgoingAnnotationSet.} This returns the outgoing annotations
of the specified anchor.  The outgoing annotations of anchor \smtt{a} are
the annotations which start with anchor \smtt{a}:

\begin{sv}
AnnotationIds GetOutgoingAnnotationSet(
                AnchorId anchorId);
\end{sv}

\begin{figure}[htbp]
  \begin{center}
    \leavevmode
    \epsfig{figure=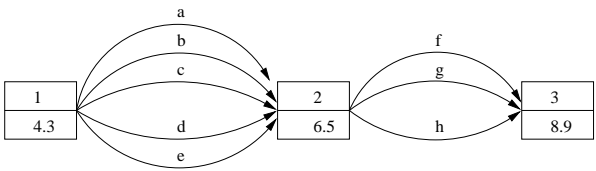,width=0.95\linewidth}  
  \end{center}
  \caption{Incoming and outgoing annotations}
  \label{fig:incoming}
\end{figure}

For example, in the annotation graph shown in Figure~\ref{fig:incoming}, the incoming
annotations of anchor 2 are a,b,c,d,e, and the outgoing annotations of
anchor 2 are f,g,h.

\noindent
\textbf{GetAnnotationSetByOffset.} This returns all annotations that
overlap a particular offset.

\begin{sv}
AnnotationIds GetAnnotationSetByOffset(
                AGId agId, Offset offset);
\end{sv}

\noindent
\textbf{GetAnnotationSeqByOffset.} This returns all annotations sorted by
start anchor offsets as the first sorting key, end anchor offsets as the
second, and \smtt{AnnotationIds} as the third.

\begin{sv}
AnnotationIds GetAnnotationSeqByOffset(AGId agId);
\end{sv}

\begin{sv}
AnnotationIds GetAnnotationSeqByOffset(
                AGId agId, Offset begin);
\end{sv}

In this case, \texttt{GetAnnotationSeqByOffset} returns all the annotations
with their start anchor offset greater than or equal to a specified offset,
sorted by start anchor offsets as the first sorting key, end anchor offsets
as the second, and \smtt{AnnotationIds} as the third.

\begin{sv}
AnnotationIds GetAnnotationSeqByOffset(
                AGId agId, Offset begin, Offset end);
\end{sv}

In this case, \texttt{GetAnnotationSeqByOffset} returns returns all the
annotations with the start anchor offset in between the specified offsets,
sorted by start anchor offsets as the first sorting key, end anchor offsets
as the second, and \smtt{AnnotationIds} as the third.

\subsubsection{XML save function}

\textbf{toXML.} This returns a string in the ATLAS Level 0 XML format of
the specified \emph{AGSet} or \emph{AG}:

\begin{sv}
string toXML(Id id);
\end{sv}

The annotation graph library also provides functions to access a
database server, for the persistent storage of annotations which
may be shared by multiple annotators, as described by
\newcite{lrec-db}.

\begin{table*}[htbp]
  \begin{center}
    \leavevmode
{\footnotesize
\begin{tabular}{|c|c|l|} \hline
Format name & Supported I/O & {Target corpus or format} \\ \hline\hline
AIF & input/output & ATLAS Interchange Format, Level 0
\myurlf{http://www.ldc.upenn.edu/AG/doc/xml/ag.dtd} \\ \hline
BAS & input & BAS Partitur format
\myurlf{http://www.phonetik.uni-muenchen.de/Bas/BasFormatseng.html} \\ \hline
BU & input & Boston University Radio Speech Corpus \\ \hline
LCF & input/output & LDC Callhome Format \\ \hline
SwitchBoard & input & Switchboard \\ \hline
TF & input/output & Table Format \\ \hline
TIMIT & input & TIMIT Corpus
\myurlf{http://www.ldc.upenn.edu/lol/docs/TIMIT.html} \\ \hline
TreeBank & input/output & Penn Treebank
\myurlf{http://www.cis.upenn.edu/~treebank/home.html} \\ \hline
xlabel & input & xlabel format \\ \hline
\end{tabular}
}
  \end{center}
  \caption{Supported file formats}
  \label{tab:formats}
\end{table*}

\section{The File I/O Library}
\label{sec:io}

The I/O library is also implemented in C++.  This section
describes how the I/O classes are used to read native format files
into annotation graphs and write them back.

Table~\ref{tab:formats} summarizes the formats which are currently
supported by the I/O library.

There is one abstract class named \smtt{agfio} which declares
the interface for the \smtt{load} and \smtt{store} methods, which are
virtual functions.  By inheriting the \smtt{agfio} class and implementing
the \smtt{load} and \smtt{store} methods, each format will be a class
with \smtt{load} and \smtt{store} methods. Loading or storing is done by
creating an instance of a format class and calling the \smtt{load} or
\smtt{store} method with the proper arguments.

As shown in Table~\ref{tab:formats}, currently only AIF, LCF and TF formats
can be stored.

\section{Scripting Language Access to the APIs}
\label{sec:scripting}

The toolkit provides interfaces to the annotation graph libraries for
the scripting languages Tcl and Python.  To avoid having to manage
object references across the Tcl/C and Python/C language interfaces
and in the event language, all communication is via strings.  These
strings hold object identifiers, feature names and feature values.
This section describes how to access the APIs from Tcl and Python.

For a Tcl program to access annotation graph functions, it must contain
either of the following declarations:

\begin{sv}
package require ag
\end{sv}

or

\begin{sv}
load ag_tcl.so
source ag.tcl
\end{sv}

The following Tcl statement creates an annotation.

\begin{sv}
AG_CreateAnnotation $agId $a1 $a2 $ann_type
\end{sv}

Note that the function name has the prefix \texttt{AG\_}.  Newer versions
of AGTK support the Tcl namespace as follows.

\begin{sv}
AG::CreateAnnotation $agId $a1 $a2 $ann_type
\end{sv}

The argument \smtt{\$agId} is an AG identifier, and the arguments \smtt{\$a1} and
\smtt{\$a2} are Anchor identifiers. The argument \smtt{\$ann\_type} is the type of
the annotation (e.g.  ``word''). 

Similarly, for a Python program, we need to import the AG module first:

\begin{sv}
import ag
\end{sv}

The following code fragment creates an AGSet, a timeline, an AG,
two anchors and an annotation. 

\begin{sv}
agSetId = ag.CreateAGSet('Test')
timelineId = ag.CreateTimeline(agSetId)
agId = ag.CreateAG(agSetId, timelineId)
anc1 = ag.CreateAnchor(agId)
anc2 = ag.CreateAnchor(agId)
ann1 = ag.CreateAnnotation(agId, anc1, anc2, "Word")
\end{sv}

The following fragment specifies features for a particular annotation,
and prints the AIF representation to standard output.

\begin{sv}
ag.SetFeature(ann1, "English", "cat")
ag.SetFeature(ann1, "Japanese", "neko")
print ag.toXML(agId),
\end{sv}

\section{Building Annotation Tools with Tcl/Tk and Python}
\label{sec:building}

Most large-scale annotation projects must deal with multiple tools
and formats.  If all required tools were developed by a single
project, and are used exactly as the original developers intended,
there is usually no problem with interoperability.  However, in 99\%
of the remaining cases, format translation and tool interoperability
is a critical issue.

If annotation tools are created around a common architecture and a
shared data model, these issues do not arise.  New file formats
are supported by writing a converter between the the format and the
general-purpose data model.  The sharing of components among
different annotation tools is straightforward, and new applications can be
developed quickly using existing components.  New functionalities added to a
component can be used by all existing tools which include that component.

The annotation graph library and the file I/O library provide the means to
create, manipulate, read and write annotation graph data.  In this section,
we explain our approach to tool creation using these libraries as the core
of the tool architecture.

All of our annotation tools are created using scripting languages
having easy-to-use GUI libraries.  This permits a rapid development
cycle and easily customizable user interfaces.  Tools are relatively
small since much of the work is done by AGTK, and this small size
and common data model mean that tool components are readily repurposable.
We deliberately avoid the temptation to create general purpose annotation
tools since each annotation task is idiosyncratic.  Annotators work
best when they use an interface which is maximally ergonomic for the
peculiarities of the task, as compared with tools that include much irrelevant
functionality and have an interface that is balanced for a wide variety of
tasks.

This section describes the steps in creating a new special-purpose tool
based on the general-purpose components provided by AGTK.  It includes
a discussion of inter-component communication, and of an example tool.

\subsection{Building tools with Tcl/Tk}

The Tk toolkit provides a graphical user-interface for Unix, Windows and
Macintosh platforms.  Some graphical user-interface components, called
widgets, are provided with the Tcl/Tk standard distribution.  For example
the following fragment of Tcl/Tk code creates and displays a text widget:

\begin{sv}
set t [text .t]
pack \$t
\end{sv}

Other GUI components, such as buttons, menubars and canvases, come standard
with Tk.  In addition, open-source GUI components and extensions are
developed and distributed by various developers around the world.
Provides pointers to such software are provided at
[\smtt{http://dev.scriptics.com}].

\subsection{Building tools with Python}
\label{sec:build:python}

Python has become very popular as a scripting language.  Python provides an
object-oriented programming framework, and is easy to learn.  A GUI package
called Tkinter \cite{Grayson00} based on the Tk toolkit is included in the
standard Python distribution. Tkinter provides class definitions for the
standard widgets included in Tk.  In addition, it is relatively easy to
write a Tkinter class for a third-party Tk widget that is not directly
supported by Tkinter.

The following segment of Python code creates and displays a Tk text
widget. 

\begin{sv}
from Tkinter import *
root = Tk()
t = Text(root)
t.pack
root.mainloop()
\end{sv}

\subsection{The inter-component communication model}
\label{sec:build:message}

An annotation tool built with the Annotation Graph Toolkit will consist of several major
components, including:
(i) a main program (script);
(ii) an annotation/transcription component in which the user
would enter annotations and transcriptions, and
(iii) a signal display component giving
access to recorded digital signals, such as speech waveforms. 
Typically, annotation/transcription components and waveform display
components are reused by different specialized annotation tools.  To create
a new annotation tool, the developer writes a main program using the
right selection of widgets and provides callback functions to handle
widget events.

\begin{figure}[htbp]
  \begin{center}
    \leavevmode
  \epsfig{figure=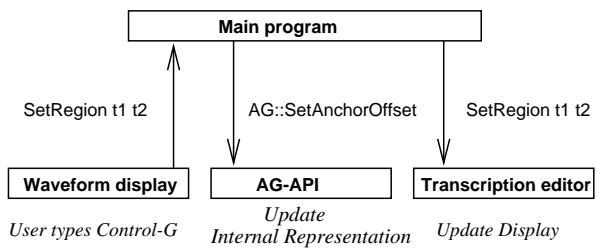,width=0.95\linewidth}  
  \end{center}
\caption{An example of inter-component message passing}
\label{fig:inter}
\end{figure}

Events are passed around among components so that necessary tasks can be
performed within each component.
Consider the following example.  Suppose that the user already has an
annotation assigned to a specific region in the signal.  He/she now wants
to assign new start and end offsets for the signal to the annotation.
Suppose the keyboard input \emph{Control-g}, in the waveform component is
assigned to such a task.  When the user hits \emph{Control-g} while there
is a newly highlighted region in the waveform, this information
(\emph{event}) needs to be passed to the main program, and then to the
annotation/transcription component and the annotation graph library.
This propagation of event information is illustrated in
Figure~\ref{fig:inter}.

The event \emph{SetRegion} is generated in the waveform component and
passed to the main program with the two parameters, the start time and the
end time.  Then, the main program sends \emph{SetRegion} to the
transcription component so that the start and end offsets can be updated in
the transcription component.  Also, the main program uses
the annotation graph function \emph{SetAnchorOffset} to update the internal
representation of the annotation graph data.
Table~\ref{tab:events} shows a list of typical events passed around in this
manner.

\begin{table}[htbp]
  \begin{center}
    \leavevmode
\begin{tabular}{|c|c|} 
\hline
Event name & Typical parameters \\
\hline
CreateAnnotation & start time, end time\\
DeleteAnnotation & annotation identifier\\
SetFeature & feature, value\\
SetRegion & start time, end time\\
GetRegion & start time, end time\\
SetCurrentAnnotation & annotation identifier\\
Play & start time, end time\\
Stop & \\
\hline
\end{tabular}
  \end{center}

  \caption{A list of common events}
  \label{tab:events}
\end{table}

These events and their necessary arguments are passed as associative
arrays (e.g., Tcl arrays and Python dictionaries).  These arrays are passed
to event handlers defined in the recipients.

\subsection{GUI design}

A critical aspect of designing a successful annotation tool is to make it
highly ergonomic for the particular annotation task.  Writing a tool in a scripting 
language makes it easy to experiment and change various aspects of a user
interface, and facilitates ``power users'' who can tweak the code to help
streamline their work.

\begin{figure}[htbp]
  \begin{center}
    \leavevmode
  \epsfig{figure=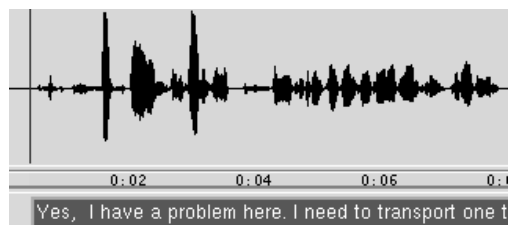,width=0.80\linewidth}  
  \end{center}
\caption{Before Split}
\label{fig:split2}
\end{figure}

\begin{figure}[htbp]
  \begin{center}
    \leavevmode
  \epsfig{figure=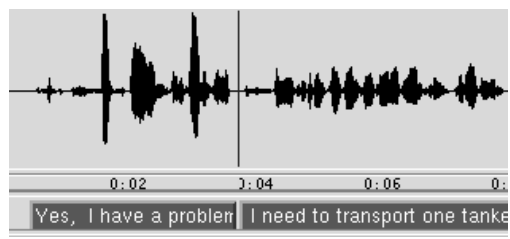,width=0.80\linewidth}  
  \end{center}
\caption{After Split}
\label{fig:split4}
\end{figure}

Where possible, our tools use simple keybindings that work in multiple
contexts, making it easy for users to focus on the annotation task instead
of having to hunt for obscure key combinations.  In the
MultiTrans annotation tool, a tool based on AGTK, the \emph{Return} key
is active in multiple contexts.  If a new region in the waveform
display is chosen (i.e., highlighted), pressing the \emph{Return} key will
create and insert a new annotation into the transcription display.  If an
existing annotation in the transcription display is chosen, its
corresponding region in the waveform is highlighted.  A single mouse-button
click in the highlighted region will set a point. If the \emph{Return} key
is pressed then, it will split the annotation and the region.  
Figure~\ref{fig:split2} and Figure~\ref{fig:split4} show before and after
split, respectively.

\begin{figure}[htbp]
  \begin{center}
    \leavevmode
  \epsfig{figure=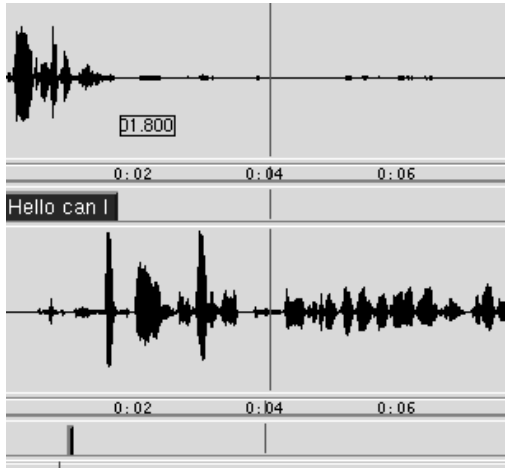,width=0.80\linewidth}  
  \end{center}
\caption{Setting Anchor Point during Playback}
\label{fig:nonspe}
\end{figure}

If the \emph{Return} key is pressed during a playback, it will set an
anchor point.  When the playback of speech is begun, the user may press the
\emph{Return} key to insert an anchor in the current channel. When
\emph{Return} is pressed a small black bar will appear below the waveform
(Figure~\ref{fig:nonspe}).  This designates the current starting position of
the annotation. When \emph{Return} is pressed a second time the end anchor
for the annotation is inserted and the annotation is created.

All these different actions have a common theme, namely to ``create''
something.  This heavy re-use of a keybinding requires a little more
effort in the coding process, but we have found that it makes the
annotation process faster and easier.  Using scripting languages,
it is very easy to experiment with the user interface in this way.

\subsection{Audio and video display using 
third-party software: WaveSurfer and QuickTime}

WaveSurfer \cite{Sjolander00} was developed by K\r{a}re Sj\"{o}lander and
Jonas Beskow of KTH as a tool for displaying and manipulating sound files.
WaveSurfer uses Snack \cite{Snack} as its signal processing module.  Both
packages are distributed under an open source license.
WaveSurfer is written in Tcl/Tk and its widget,
called \emph{wsurf}, can be embedded in an application written in Tcl/Tk.
A Python interface has also been developed.  This makes WaveSurfer an
excellent component to use with AGTK.

QuickTime Tcl is another component that can be used with AGTK.
QuickTime Tcl requires the QuickTime player created by Apple Corporation.
Currently, no UNIX version of QuickTime player is provided by Apple.

\subsection{Embedding a third-party Tk widget in Python}

Tkinter provides class definitions only for the original Tk widgets.
However, it is relatively easy to write an extension of Tkinter for a
third-party Tk widget.  For example, AGTK includes a Python/Tkinter class
definition for the Wsurf widget.  The class \emph{Wsurf} covers the Wsurf
API.  The class \emph{agWsurf}, which is a subclass of \emph{Wsurf},
provides methods specific to the tasks required by the tools we create.



\subsection{A Tk table widget}
\label{sec:build:table}

AGTK comes with a table annotation component called agTable (or,
ag-table as originally called in the Tcl version).  The agTable component
is based on TkTable written by Jeffrey Hobbs, et.\ al.
[\smtt{http://sourceforge.net/projects/tktable/}].








%

\subsection{Putting it all together: the case of TableTrans}

\begin{figure}[htbp]
  \begin{center}
    \leavevmode
  \epsfig{figure=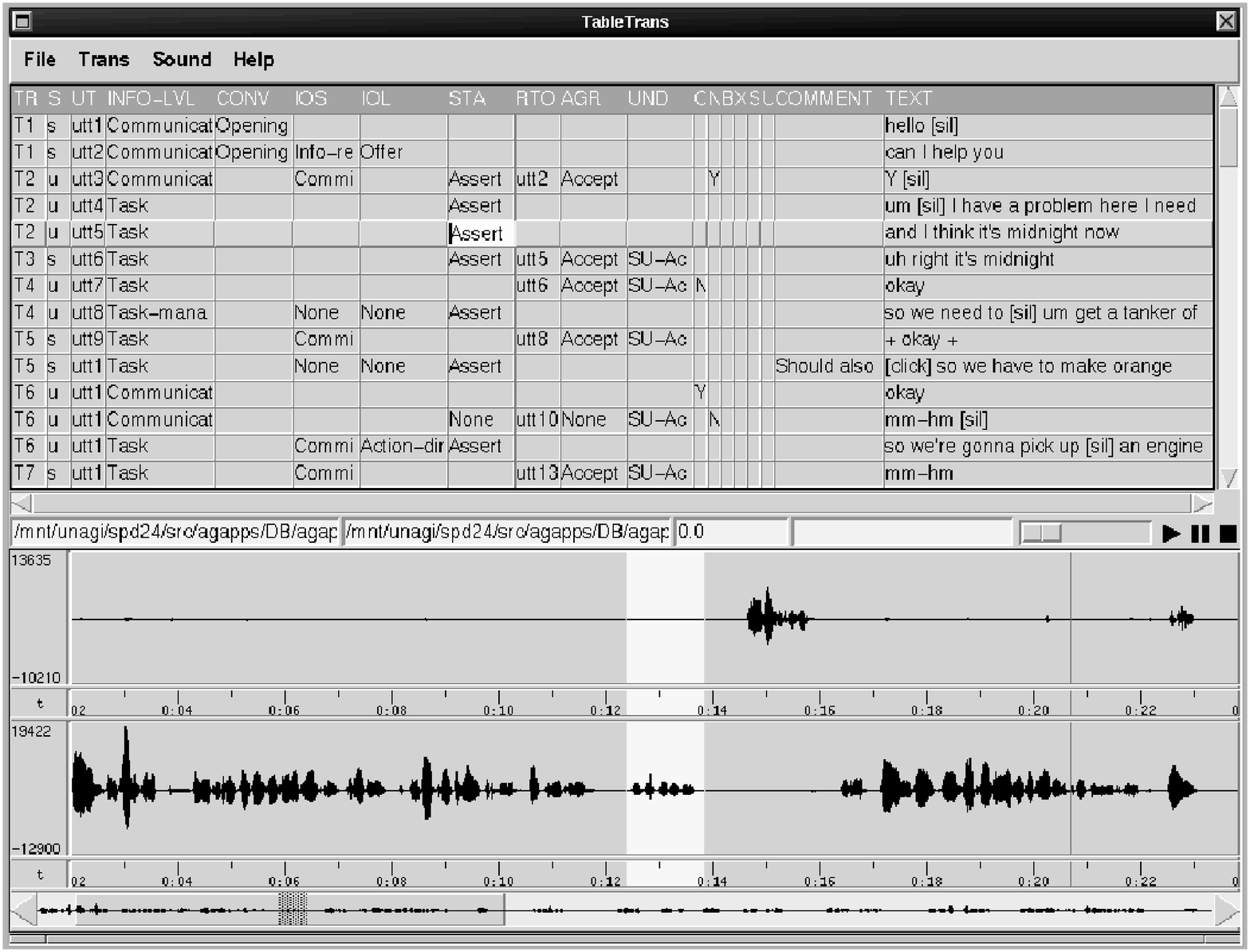,width=1.00\linewidth}
  \end{center}
\caption{TableTrans}
\label{fig:tabletrans}
\end{figure}

Now that we have seen the major components of AGTK, we can examine how to
build an annotation tool using these components.  As an example, we will
examine an annotation tool called TableTrans (Figure~\ref{fig:tabletrans}).
TableTrans is a spreadsheet-style annotation tool that
uses the components we have described in this paper.

TableTrans consists of three major components:
(i) the main script, \emph{agTableTrans}, 
(ii) the table component, \emph{agTable}, and
(iii) the waveform display component, \emph{agWsurf}.
Here we will look at two common operations:
pressing the \emph{Return} key in the table to insert a new annotation,
and pressing the \emph{Control-d} key combination in the table to delete
the highlighted annotation.

The keybindings for \emph{Return} and \emph{Control-d} are already defined
in the agTable component.  The operations above pass the events
\emph{CreateAnnotation} and \emph{DeleteAnnotation} to the main script,
respectively. 
A callback function (\emph{agTableEvent}) is defined in the main
script.  When the \emph{Return} key is pressed, the table component 
sends the event \emph{CreateAnnotation} to the main script.  The callback
function in the main script then performs the following tasks:

First, using the start and end offsets of the current region stored in the
main script, it calls the annotation graph function \emph{CreateAnchor} twice to create
the start anchor and the end anchor:

\begin{sv}
a1 = ag.CreateAnchor(self._AGName)
ag.SetAnchorOffset(a1, self._currentStartPosition)
a2 = ag.CreateAnchor(self._AGName)
ag.SetAnchorOffset(a2, self._currentEndPosition)
\end{sv}

Second, it calls the annotation graph function \emph{CreateAnnotation} using the
start and end anchors:

\begin{sv}
id = ag.CreateAnnotation(self._AGName,
                            a1, a2, annotationType)
\end{sv}

This function returns the annotation identifier.  Finally, the program
returns the new identifier, along with the start and end offsets, to the
table component.  The table component inserts a new row using the
annotation identifier and the start and end offsets.

Similarly when the \emph{Control-d} key combination is pressed, the table
component sends the event \emph{DeleteAnnotation} to the main script
together with the annotation identifier of the highlighted row.  The callback
function in the main script performs the following command, where
dictionary \emph{event} contains the event message.

\begin{sv}
ag.DeleteAnnotation(event['AnnotationId'])
\end{sv}

\begin{figure}[htbp]
  \begin{center}
    \leavevmode
  \epsfig{figure=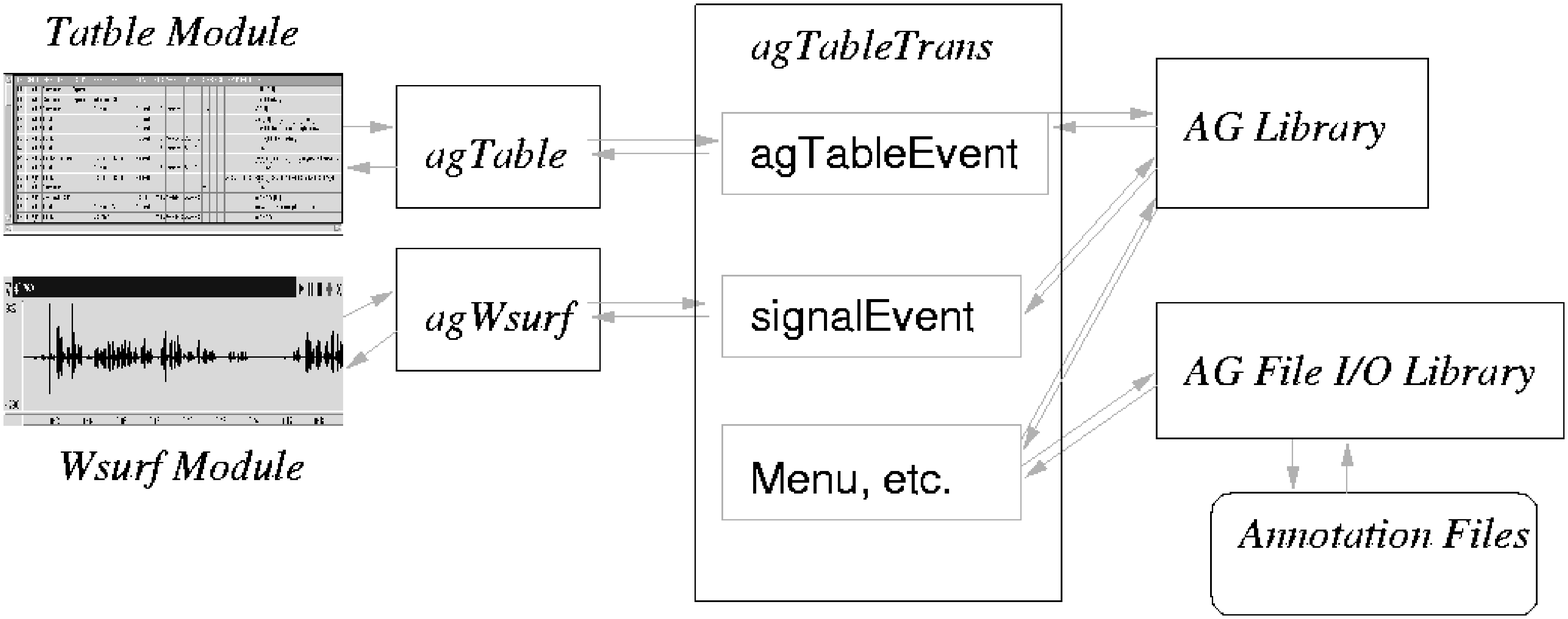,width=1.00\linewidth}
  \end{center}
\caption{Components in TableTrans}
\label{fig:ttdiagram}
\end{figure}

Then, the table component can remove the highlighted row from its table.
Figure~\ref{fig:ttdiagram} shows the components of TableTrans and their
relationships.



\section{Conclusion}

This paper has described a new toolkit, AGTK, which supports rapid
development of linguistic annotation software.
AGTK is being used for annotation projects at the Linguistic Data
Consortium \myurl{www.ldc.upenn.edu}.
The toolkit is available for download at
\myurl{http://www.sourceforge.net/projects/agtk/}.  For news and updates,
please visit \myurl{http://www.ldc.upenn.edu/AG/}.

Existing third-party tools, namely Emu and Transcriber
\cite{CassidyHarrington01,Barras01}, are being migrated to AGTK.
They will share the same internal data model and relational storage model,
while keeping their distinctive user interfaces and file formats.  Once
these ports have been completed, we will have a shared library of user
interfaces to complement the AG and file I/O libraries.
We hope that these shared libraries will continue to grow as members of the
wider community contribute I/O and GUI components to AGTK.

In future work, we hope to develop more applications for annotation in
other areas, possibly including: sociolinguistics, conversational analysis,
sign and gesture, discourse and dialogue.  On the
technical side, we hope to add interfaces to video widgets on all
platforms, and to support data entry for extended-Roman and non-Roman scripts.

\section*{Acknowledgements}

This material is based upon work supported by the National Science
Foundation under Grant Nos. 9978056, 9980009 (Talkbank).  The authors
are grateful to Claude Barras, Steve Cassidy, Mark Liberman and many
others for discussions on the material presented here.



\bibliographystyle{lrec2000k}

\end{document}